\documentclass[10pt,twocolumn]{article}

\usepackage[margin=0.75in]{geometry}
\usepackage{times}
\usepackage{graphicx}
\usepackage{amsmath}
\usepackage{amssymb}
\usepackage{booktabs}
\usepackage{array}
\usepackage{url}
\usepackage{eso-pic}
\usepackage[hidelinks]{hyperref}

\title{Does the Judge Prefer English?\\
Evaluating Language-Switching Invariance in LLM-as-a-Judge}

\author{Shaojie Yin\\
Shanghai Artificial Intelligence Laboratory\\
Shanghai, China\\
\texttt{Misaka10086@sjtu.edu.cn}\\
\href{https://orcid.org/0009-0008-1526-1824}{ORCID: 0009-0008-1526-1824}}
\date{}

\begin{document}
\AddToShipoutPictureFG*{%
  \put(430,735){\includegraphics[width=1.55in]{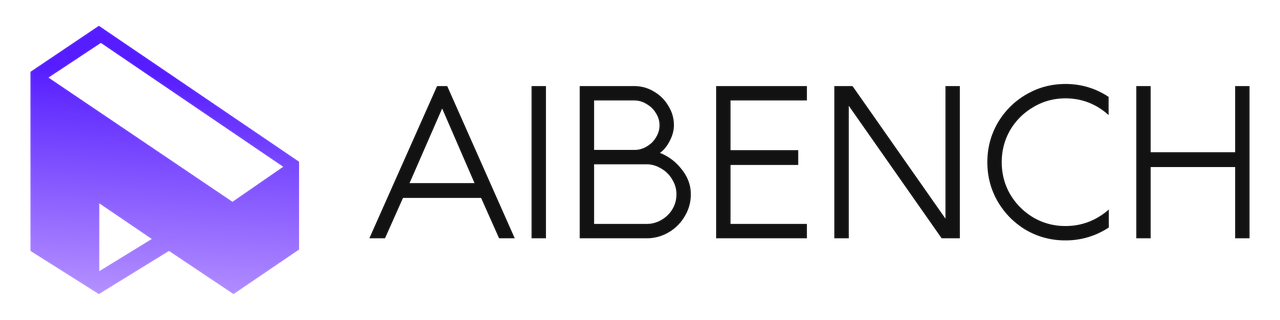}}%
}
\maketitle

\begin{abstract}
Large language models (LLMs) are now widely used as automatic judges for open-ended instruction-following evaluation. This practice is convenient, scalable, and often more semantically aware than reference-based metrics, but it also introduces a new reliability question: does a judge evaluate the quality of an answer, or does it also react to the language in which the comparison is presented? We propose Judge-LS, a lightweight meta-evaluation protocol that transforms LLMBar response-pair items into English, Chinese, and Chinese-English language-switched variants. A reliable judge should preserve its preference under label-preserving language transformations and should not prefer a language when two answers are translation-equivalent. We evaluate four API-accessible judges on the full 419-item LLMBar benchmark, producing 13,408 successful pairwise judgments. Across models, Chinese and language-switched presentations induce 10.7--14.4\% preference flips relative to English, and all judges achieve their highest accuracy in English. However, translation-equivalent tie probes do not reveal a systematic English preference: most probes are judged as ties, and non-tie decisions more often favor Chinese. We add confidence intervals, paired significance tests, and an automatic transformation audit with a sensitivity analysis that excludes mechanically flagged high-risk variants. The experiment requires no model training, uses only API calls, and is feasible on modest local hardware.
\end{abstract}

\section{Introduction}

Automatic evaluation has become one of the practical bottlenecks in building and comparing LLM systems. Human preference annotation is expensive, slow, and difficult to repeat whenever a model, prompt, or decoding strategy changes. As a result, LLM-as-a-Judge has become a common replacement or pre-filter for human evaluation. A judge model is given an instruction and one or more candidate answers, and is asked to produce a preference, score, or explanation. This setting has been used for chat benchmarks, instruction-following comparisons, and increasingly for multimodal or domain-specific evaluation.

The appeal of LLM judges is clear: they can read open-ended outputs, reason over task requirements, and produce structured decisions at large scale. The risk is also clear: the judge itself is another model with its own biases. Prior work has documented position bias, verbosity bias, self-preference, prompt sensitivity, and limitations in judging instruction-following behavior \cite{liu2023geval,zeng2024llmbar,zheng2023judging,dubois2024lengthcontrolled,wang2024fair,panickssery2024selfpreference}. These biases matter because evaluation results influence model selection, leaderboard rankings, and research conclusions.

This paper studies a smaller but important reliability question: \emph{does a judge preserve its preference when the same pairwise comparison is presented in English, Chinese, or Chinese-English language-switching?} If two answer pairs differ only by a label-preserving language transformation, a judge should ideally make the same choice. If the best answer is translated into Chinese and compared against the same English answer, the judge should usually call a tie rather than reward a particular language. Violations of these expectations imply that the evaluation pipeline is sensitive to surface presentation, not only response quality.

We focus on English--Chinese because Chinese is widely used in real multilingual LLM deployments and because language-switching is common in technical, educational, and online communication. The study is designed for low-resource experimental settings: no model is trained, no local large model inference is required, and a modest local GPU is not a bottleneck. All generation and judging are done through API calls, while the local machine performs orchestration, parsing, plotting, and LaTeX compilation.

Our contributions are:
\begin{itemize}
    \item We introduce \textbf{Judge-LS}, a simple protocol for testing language-switching invariance in pairwise LLM-as-a-Judge evaluation.
    \item We transform the full 419-item LLMBar benchmark into English, Chinese, and Chinese-English language-switched variants, preserving the original gold preference labels.
    \item We evaluate four API-accessible judges using 13,408 successful judgments, including answer-order swaps and translation-equivalence tie probes.
    \item We report not only accuracy, but also language-invariance flip rate, gold-correctness flip rate, position inconsistency, source-level robustness, language preference, uncertainty estimates, paired significance tests, and usage cost.
\end{itemize}

\begin{figure*}[t]
\centering
\includegraphics[width=\textwidth]{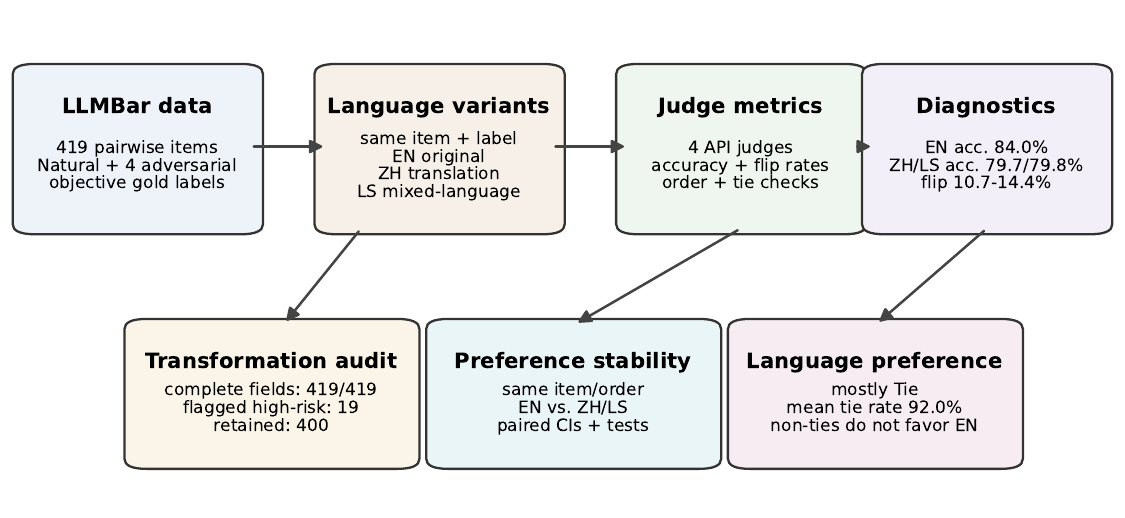}
\caption{Overview of Judge-LS. The full LLMBar response-pair benchmark is converted into English, Chinese, and Chinese-English language-switched variants. Four API judges are evaluated with answer-order swaps and translation-equivalent tie probes, then summarized by accuracy, judgment-flip rate, correctness-flip rate, position inconsistency, language preference, and transformation-audit diagnostics.}
\label{fig:overview}
\end{figure*}

\section{Related Work}

\paragraph{Automatic text evaluation.}
Early automatic evaluation metrics made large-scale model comparison practical by replacing or reducing human annotation. BLEU and ROUGE use lexical overlap for machine translation and summarization evaluation \cite{papineni2002bleu,lin2004rouge}. Later learned or embedding-based metrics such as BERTScore and COMET improve semantic matching and multilingual machine-translation quality estimation \cite{zhang2020bertscore,rei2020comet}. LLM-as-a-Judge continues this trajectory toward semantic evaluation, but it also inherits a harder reliability problem: the evaluator is a generative model whose decisions can change with prompt form, answer order, or language presentation.

\paragraph{LLM-as-a-Judge.}
G-Eval showed that GPT-4-style evaluators can align better with human ratings than many traditional natural language generation metrics when prompted with explicit criteria \cite{liu2023geval}. MT-Bench and Chatbot Arena popularized LLM judges for pairwise conversational evaluation and analyzed biases such as position and verbosity effects \cite{zheng2023judging}. AlpacaFarm and AlpacaEval use low-cost pairwise feedback or auto-annotation to compare instruction-following models, while length-controlled AlpacaEval explicitly debiases verbosity effects \cite{dubois2023alpacafarm,dubois2024lengthcontrolled}. Prometheus studies open evaluator models trained for fine-grained rubric-based assessment \cite{kim2024prometheus}. LLMBar targets a stricter meta-evaluation setting: an evaluator must identify which of two outputs objectively follows an instruction better \cite{zeng2024llmbar}. We use LLMBar because its gold labels make it possible to measure when a judge's decision changes under a language transformation.

\paragraph{Judge bias and calibration.}
FairEval shows that simply swapping candidate order can alter LLM-judge rankings and proposes position-balancing calibration \cite{wang2024fair}. Other work shows that automatic evaluators may favor longer answers or even their own generations \cite{dubois2024lengthcontrolled,panickssery2024selfpreference}. These studies motivate our answer-order swaps and our decision to report invariance metrics rather than only aggregate accuracy. Judge-LS adds a language-surface intervention: the answer content is intended to remain label-preserving, while the presentation language changes.

\paragraph{Multilingual judge reliability.}
Multilingual LLM evaluation is not just English evaluation translated into other languages. A judge may have different competence, calibration, and cultural assumptions across languages. Broad multilingual benchmarks such as XTREME, MEGA, and multilingual ChatGPT evaluations show that model behavior varies substantially across languages and resource levels \cite{hu2020xtreme,bang2023multitask,lai2023chatgpt,ahuja2023mega}. Recent work on multilingual LLM-as-a-Judge reliability reports low agreement in many non-English settings \cite{fu2025reliable}. Judge-LS is complementary: instead of surveying many languages, it isolates a controlled English--Chinese--language-switching intervention on the same pairwise comparisons. This lets us ask whether the judge is invariant to language presentation when the intended quality relation is fixed.

\paragraph{Evaluation benchmarks and domain robustness.}
Judge-LS is also motivated by broader evaluation work from trustworthy AI and multimodal benchmarking. AIBench studies trustworthy evaluation principles under the 45$^\circ$ law, while a recent large multimodal model survey summarizes evaluation dimensions for modern multimodal systems \cite{aibench2025,zhang2025large}. Related first-author work by Tian studies video understanding, action recognition, video compression, and representation learning benchmarks, including early ASD detection, event-adaptive action recognition, low-bitrate video understanding, semantic video compression, video rescaling, and self-supervised motion representation \cite{tian2019asd,tian2022ean,tian2024lowbitrate,tian2023nonsemantics,tian2023clsa,tian2021selfconditioned,tian2020motion,tian2025smcpp}. These efforts motivate our emphasis on evaluator robustness rather than treating benchmark scores as transparent measurements.

\section{Judge-LS Protocol}

\subsection{Base Data}

We use LLMBar, a 419-item benchmark for evaluating whether an LLM evaluator can judge instruction-following responses \cite{zeng2024llmbar}. Each item contains an instruction, two candidate outputs, and an objective gold label indicating whether output 1 or output 2 better follows the instruction. LLMBar contains one natural subset and four adversarial subsets. In our full experiment, the source distribution is:

\begin{center}
\small
\begin{tabular}{lr}
\toprule
Subset & Items \\
\midrule
Natural & 100 \\
Adv-Neighbor & 134 \\
Adv-GPTInst & 92 \\
Adv-GPTOut & 47 \\
Adv-Manual & 46 \\
\bottomrule
\end{tabular}
\end{center}

Using all LLMBar items avoids the sampling uncertainty of the earlier pilot run. Because the benchmark is pairwise and label-bearing, it supports both standard accuracy and invariance diagnostics.

\subsection{Language Transformations}

For every LLMBar item we keep the original English version and create two additional versions:

\begin{itemize}
    \item \textbf{ZH}: the instruction and both candidate outputs are translated into natural Simplified Chinese.
    \item \textbf{LS}: the item is converted into natural Chinese-English language-switching, preserving English task terms, named entities, or technical terms when appropriate.
\end{itemize}

We use \texttt{gpt-4.1-mini} as the transformation model. The prompt explicitly asks the model to preserve every factual, mathematical, formatting, and instruction-following error. This is important because a translation that silently repairs the weaker answer would invalidate the original gold label. One item repeatedly produced invalid JSON because of nested quotation marks in a vocabulary quiz; we manually repaired that item while preserving the same response-quality relation.

\subsection{Transformation Audit}

Because label preservation is a central assumption, we add an automatic audit before interpreting judge behavior. The audit verifies that every unique item has complete English, Chinese, and language-switched instruction and response fields, then conservatively flags variants that may need human review because of empty fields, severe length shrinkage, or large numeric-token count changes. These checks are mechanical rather than semantic: for example, a numeric-token mismatch can be caused by a valid Chinese unit conversion. We therefore keep the full 419-item set as the primary experiment and report a sensitivity analysis that excludes the 19 flagged high-risk items.

\subsection{Judgment Construction}

For each item and each language condition, the judge receives the instruction and two responses labeled A and B. We evaluate both original and swapped response order. After the model chooses A, B, or Tie, we map the output back to the original output IDs. This produces six quality judgments per item per judge:

\[
3 \text{ language conditions} \times 2 \text{ answer orders}.
\]

We also create two tie probes per item. The gold answer in English is compared with its Chinese translation in both orders. In this setting, the intended answer is \textsc{Tie}. We use the gold answer because it is the response most likely to be semantically complete and least likely to contain fragile errors whose translation could create a new quality difference. A weaker-answer tie probe would be useful, but it can confound language preference with error-preservation drift. A judge that chooses English or Chinese in many gold-answer probes is therefore exhibiting language preference rather than quality sensitivity under a conservative high-quality equivalence condition. Each item yields eight judgments per model.

\section{Metrics}

\paragraph{Tie-half accuracy.}
Let $\hat{y}_j$ be the normalized judgment for quality comparison $j$ and $y_j$ be the LLMBar gold label. We report strict accuracy and tie-half accuracy:

\[
\mathrm{Acc}_{1/2} =
\frac{1}{N}\sum_j
\begin{cases}
1 & \hat{y}_j = y_j,\\
0.5 & \hat{y}_j = \textsc{Tie},\\
0 & \text{otherwise}.
\end{cases}
\]

Tie-half accuracy is useful because some model outputs genuinely report uncertainty. Strict accuracy counts all ties as incorrect. We report both metrics in the paper because they answer different questions: strict accuracy measures decisive agreement with LLMBar, while tie-half accuracy avoids treating uncertainty as fully wrong.

\paragraph{Language-invariance flip rate.}
For target language $l \in \{\mathrm{ZH},\mathrm{LS}\}$, we compare the normalized judgment in English with the judgment under the transformed presentation:

\[
\mathrm{Flip}(l)=
\frac{1}{N}\sum_{i,o}\mathbb{1}
[\hat{y}_{i,\mathrm{EN},o}\neq\hat{y}_{i,l,o}],
\]

where $i$ indexes items and $o$ indexes answer order. We also report gold-correctness flip rate: whether the judge changes from correct to incorrect, or from incorrect to correct, after the language transformation.

\paragraph{Position inconsistency.}
For each item and language condition, we compare the normalized decision before and after swapping response order. If the normalized winner changes, the judge is position-inconsistent. This metric is especially important because language effects and answer-order effects can interact.

\paragraph{Tie-probe language preference.}
For translation-equivalent probes, we count English wins, Chinese wins, and ties. We also record the English share among non-tie decisions. A high English win rate would support the hypothesis that the judge prefers English; a high flip rate without English wins instead suggests instability rather than simple English preference.

\paragraph{Uncertainty and paired tests.}
For binomial rates such as flip rate, position inconsistency, and tie-probe outcomes, we report Wilson 95\% confidence intervals. For tie-half accuracy, which is an average over $\{0,0.5,1\}$ scores, we report normal 95\% intervals over per-judgment scores. To test whether language changes alter strict gold correctness, we use an exact two-sided McNemar/binomial test on paired English-vs.-transformed judgments with the same item and answer order.

\section{Experimental Setup}

We evaluate four API-accessible judges:
\begin{itemize}
    \item GPT-4.1 Mini: \texttt{gpt-4.1-mini}
    \item Claude Haiku 4.5: \texttt{claude-haiku-4-5-20251001}
    \item Gemini 2.5 Flash: \texttt{gemini-2.5-flash}
    \item DeepSeek V4 Flash: \texttt{deepseek-v4-flash}
\end{itemize}

All calls use temperature 0 and a JSON-only prompt with allowed winners \texttt{A}, \texttt{B}, or \texttt{Tie}. The prompt asks for a self-reported confidence value, but we treat confidence only as logged metadata because it is not calibrated across model families. The experiment contains 419 items, four judges, and eight judgments per item per judge, yielding 13,408 successful unique judgments. The full prompt templates, parsing rules, retry logic, and analysis code are in \texttt{scripts/judge\_cs.py}. The local machine only runs Python orchestration and plotting. A local conda environment avoids package conflicts, and API credentials are supplied through environment variables rather than written to project files.

The DeepSeek endpoint often emitted long reasoning content before the final JSON answer. We therefore used a higher output cap for DeepSeek and retried rare empty-content failures. This engineering detail matters for reproducibility: if a model spends all allowed tokens on hidden or exposed reasoning, a judge pipeline may falsely record a parser failure rather than a model decision.

\section{Main Results}

\begin{table*}[t]
\centering
\small
\setlength{\tabcolsep}{6pt}
\renewcommand{\arraystretch}{1.08}
\begin{tabular}{llrrrrrr}
\hline
Model & \shortstack{Language\\condition} & N & \shortstack{Strict\\acc. (\%)} & \shortstack{Tie-half\\acc. (\%)} & \shortstack{Tie-half\\95\% CI} & \shortstack{Position-A\\rate (\%)} & \shortstack{Tie\\rate (\%)} \\
\hline
Claude Haiku & LS & 838 & 73.9 & 74.4 & [71.5, 77.3] & 49.4 & 1.1 \\
Claude Haiku & EN & 838 & 78.5 & 78.9 & [76.1, 81.6] & 52.5 & 0.7 \\
Claude Haiku & ZH & 838 & 74.0 & 74.3 & [71.4, 77.3] & 53.0 & 0.7 \\
DeepSeek & LS & 838 & 86.5 & 88.9 & [86.9, 90.9] & 50.8 & 4.8 \\
DeepSeek & EN & 838 & 89.5 & 90.5 & [88.5, 92.4] & 51.8 & 1.9 \\
DeepSeek & ZH & 838 & 86.5 & 87.8 & [85.6, 89.9] & 53.1 & 2.5 \\
Gemini Flash & LS & 838 & 82.3 & 82.9 & [80.4, 85.5] & 59.2 & 1.2 \\
Gemini Flash & EN & 838 & 87.5 & 87.8 & [85.6, 90.0] & 54.3 & 0.6 \\
Gemini Flash & ZH & 838 & 83.3 & 83.7 & [81.2, 86.2] & 56.9 & 0.8 \\
GPT-4.1 Mini & LS & 838 & 72.0 & 73.0 & [70.1, 76.0] & 50.8 & 2.1 \\
GPT-4.1 Mini & EN & 838 & 77.9 & 78.8 & [76.0, 81.5] & 53.7 & 1.7 \\
GPT-4.1 Mini & ZH & 838 & 71.7 & 72.8 & [69.8, 75.8] & 52.6 & 2.1 \\
\hline
\end{tabular}
\caption{Quality-judgment accuracy by language condition on LLMBar. Strict accuracy counts ties as incorrect; tie-half gives ties half credit. Position-A is choosing the displayed first answer before order normalization; all rates are percentages.}
\label{tab:quality}
\end{table*}

\begin{table*}[t]
\centering
\small
\setlength{\tabcolsep}{6pt}
\renewcommand{\arraystretch}{1.08}
\begin{tabular}{llrrrrr}
\hline
Model & \shortstack{Language\\pair} & N & \shortstack{Judgment\\flip (\%)} & \shortstack{Judgment flip\\95\% CI} & \shortstack{Correctness\\flip (\%)} & \shortstack{Correctness flip\\95\% CI} \\
\hline
Claude Haiku & EN--LS & 838 & 12.3 & [10.2, 14.7] & 12.0 & [10.0, 14.4] \\
Claude Haiku & EN--ZH & 838 & 14.2 & [12.0, 16.7] & 14.1 & [11.9, 16.6] \\
DeepSeek & EN--LS & 838 & 10.7 & [8.8, 13.0] & 9.9 & [8.1, 12.1] \\
DeepSeek & EN--ZH & 838 & 11.1 & [9.2, 13.4] & 10.6 & [8.7, 12.9] \\
Gemini Flash & EN--LS & 838 & 12.2 & [10.1, 14.6] & 12.0 & [10.0, 14.4] \\
Gemini Flash & EN--ZH & 838 & 11.1 & [9.2, 13.4] & 10.9 & [8.9, 13.2] \\
GPT-4.1 Mini & EN--LS & 838 & 12.9 & [10.8, 15.3] & 12.2 & [10.1, 14.6] \\
GPT-4.1 Mini & EN--ZH & 838 & 14.4 & [12.2, 17.0] & 14.1 & [11.9, 16.6] \\
\hline
\end{tabular}
\caption{Language-invariance flip rates relative to English. Judgment flip counts any changed normalized preference; correctness flip counts a change between strict gold-correct and not strict gold-correct. All rates and intervals are percentages.}
\label{tab:invariance}
\end{table*}

\begin{table}[t]
\centering
\small
\setlength{\tabcolsep}{4pt}
\renewcommand{\arraystretch}{1.08}
\begin{tabular}{lrrrrr}
\hline
Model & N & \shortstack{EN wins\\(\%)} & \shortstack{ZH wins\\(\%)} & \shortstack{Tie\\(\%)} & \shortstack{EN share\\non-tie (\%)} \\
\hline
Claude Haiku & 838 & 2.6 & 15.0 & 82.3 & 14.9 \\
DeepSeek & 838 & 0.7 & 1.8 & 97.5 & 28.6 \\
Gemini Flash & 838 & 3.1 & 4.3 & 92.6 & 41.9 \\
GPT-4.1 Mini & 838 & 0.7 & 3.7 & 95.6 & 16.2 \\
\hline
\end{tabular}
\caption{Language preference in translation-equivalent tie probes. EN/ZH win rates indicate which language version is preferred; EN share is computed over non-tie decisions. All rates are percentages.}
\label{tab:tieprobe}
\end{table}

\begin{table}[t]
\centering
\small
\setlength{\tabcolsep}{4pt}
\renewcommand{\arraystretch}{1.08}
\begin{tabular}{lrrr}
\hline
Model & Calls & \shortstack{Input\\tokens} & \shortstack{Output\\tokens} \\
\hline
Claude Haiku & 3352 & 1,431,634 & 413,642 \\
DeepSeek & 3352 & 1,104,204 & 1,363,065 \\
Gemini Flash & 3352 & 1,158,978 & 1,726,713 \\
GPT-4.1 Mini & 3352 & 1,191,632 & 183,183 \\
Variant gen. & 418 & 192,979 & 235,786 \\
\hline
\end{tabular}
\caption{Observed or approximated API usage. Costs depend on the endpoint's billing plan.}
\label{tab:usage}
\end{table}

\begin{figure*}[t]
\centering
\includegraphics[width=0.32\linewidth]{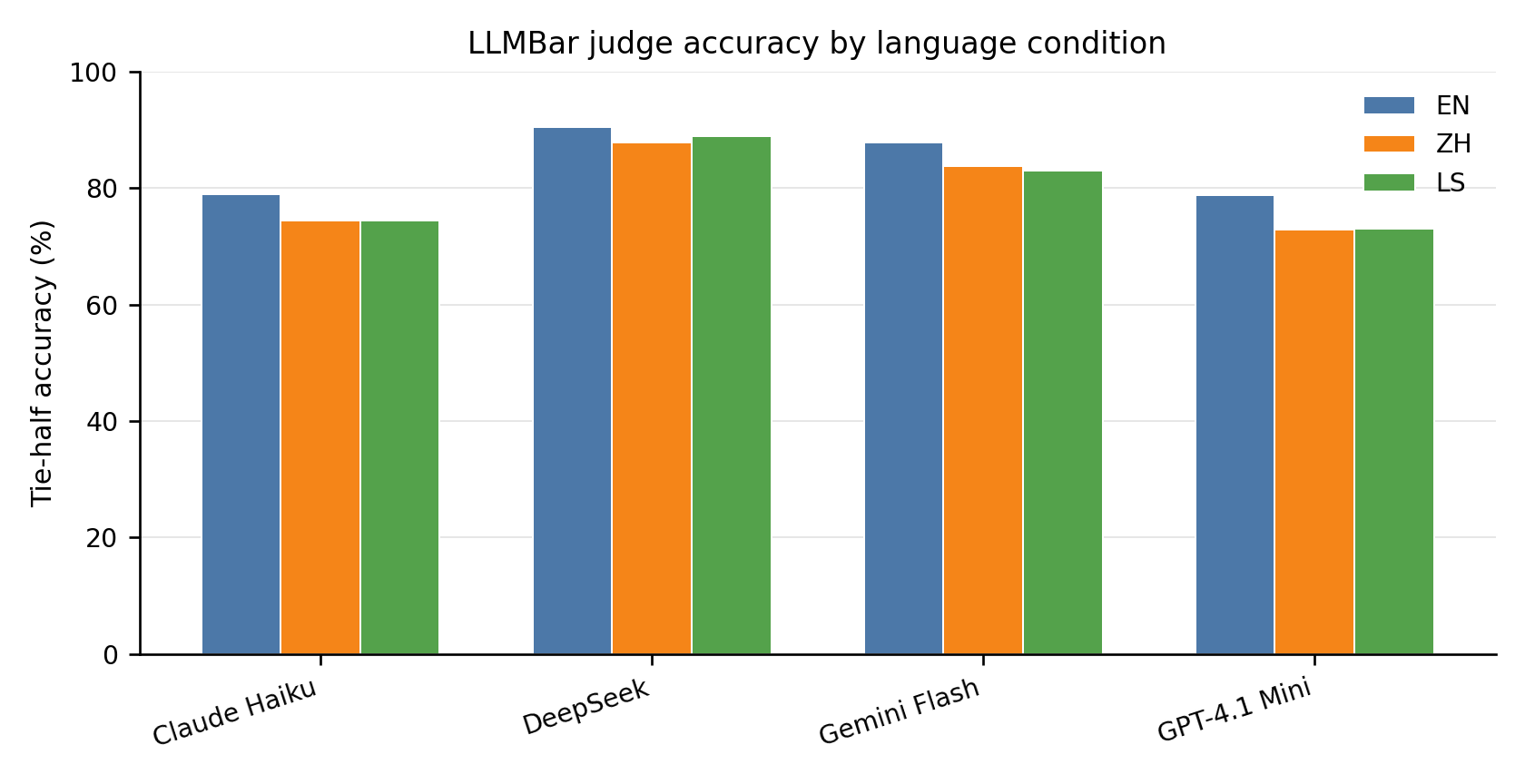}
\includegraphics[width=0.32\linewidth]{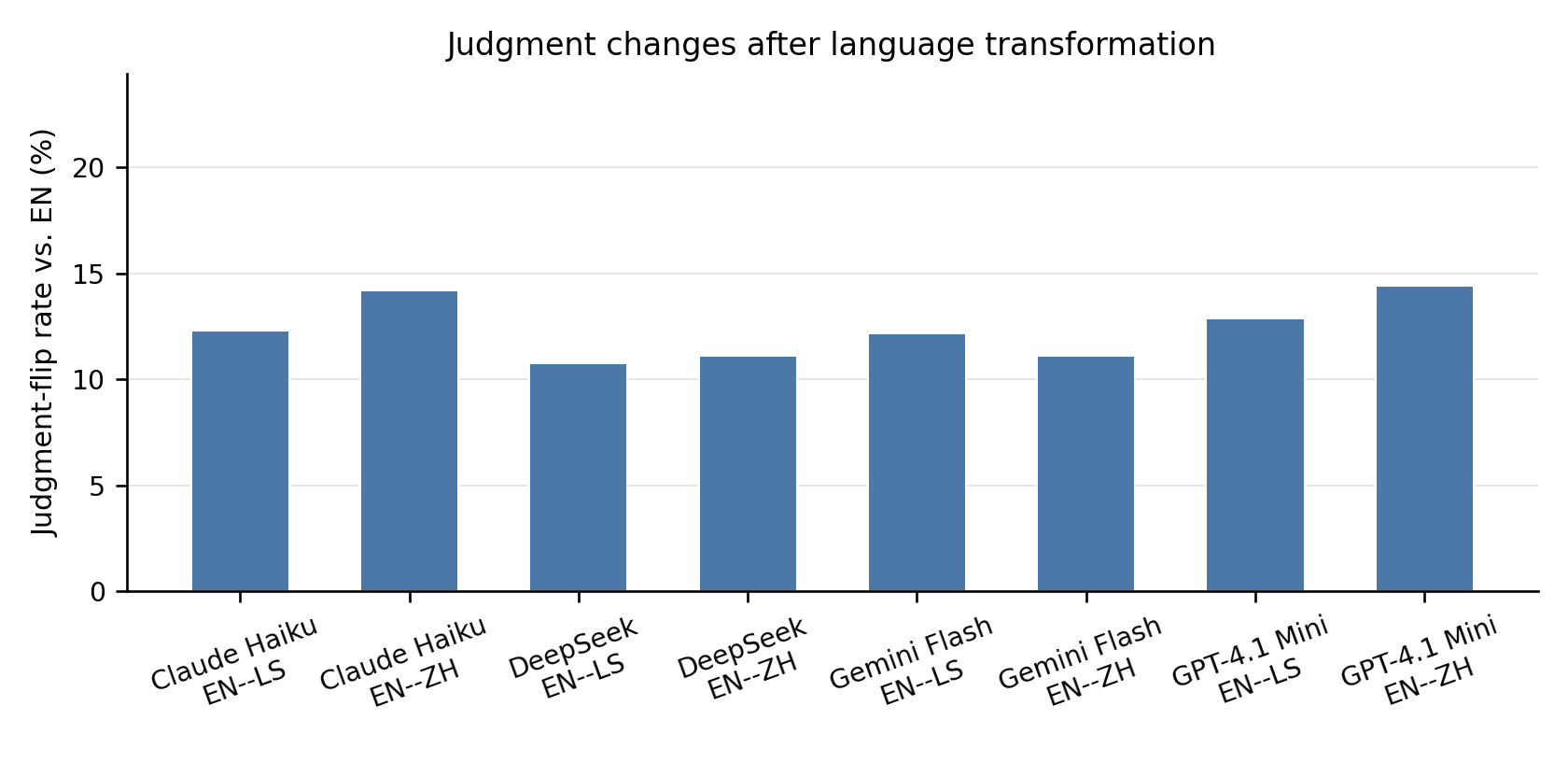}
\includegraphics[width=0.32\linewidth]{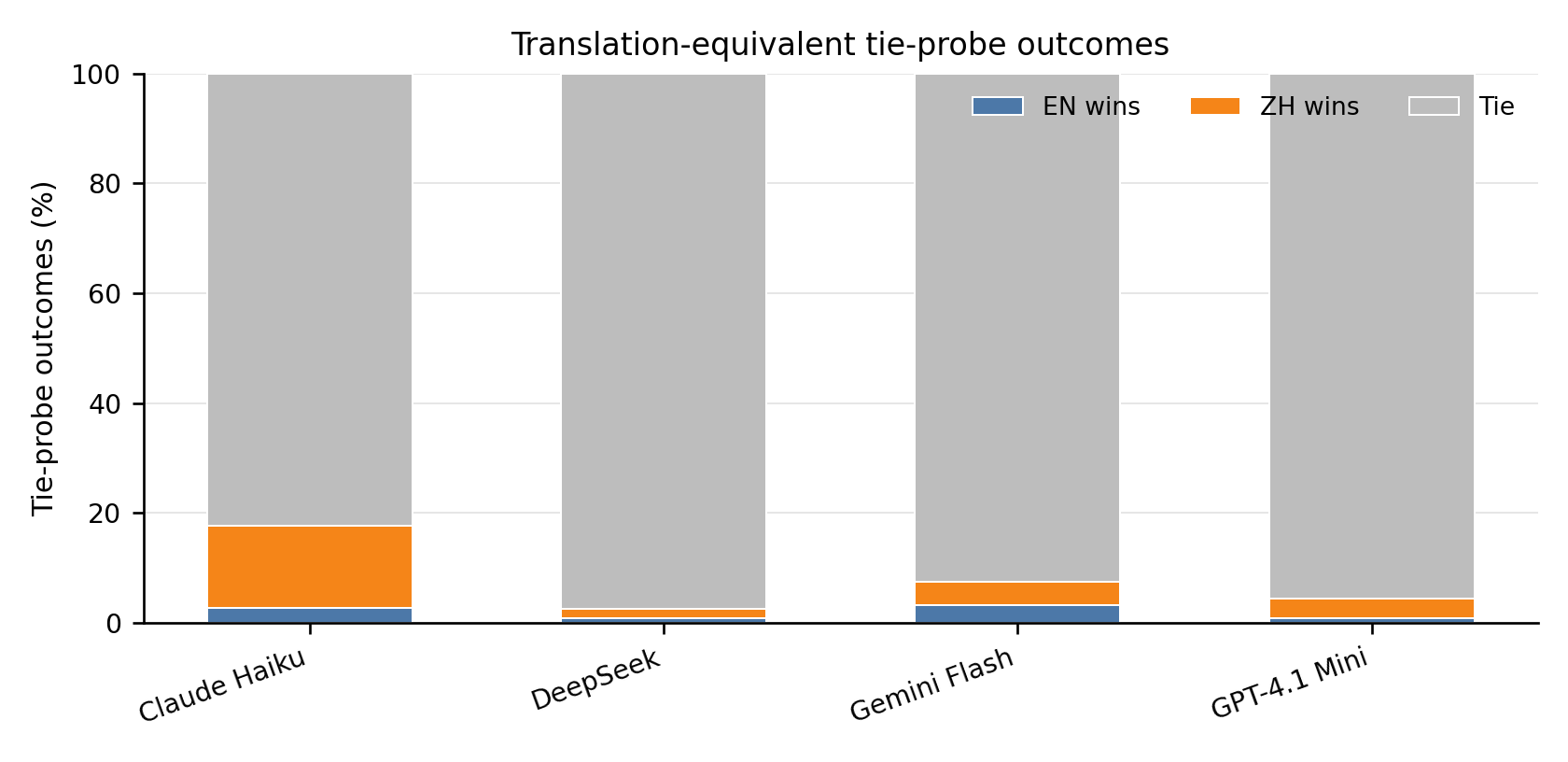}
\caption{Main results on the 419-item LLMBar experiment. Left: tie-half accuracy by language condition, with LS denoting Chinese-English language-switching. Middle: judgment-flip rate versus EN for EN--ZH and EN--LS. Right: outcomes for EN-vs.-ZH translation-equivalent tie probes.}
\label{fig:main}
\end{figure*}

\paragraph{Accuracy is highest in English.}
All four judges perform best in English (Table~\ref{tab:quality}). GPT-4.1 Mini drops from 78.8\% tie-half accuracy in English to 72.8\% in Chinese and 73.0\% in language-switched form. Claude Haiku drops from 78.9\% to roughly 74.3--74.4\%. Gemini drops from 87.8\% to 83.7\% in Chinese and 82.9\% in language-switched form. DeepSeek is strongest overall, but still drops from 90.5\% English accuracy to 87.8\% in Chinese and 88.9\% in language-switched form. The strict-accuracy column shows the same qualitative pattern, so the English advantage is not an artifact of giving half credit to ties.

\paragraph{Language transformations cause substantial preference flips.}
Language-invariance flip rates range from 10.7\% to 14.4\% (Table~\ref{tab:invariance}). GPT-4.1 Mini and Claude Haiku are most sensitive to Chinese transformations, with English-vs.-Chinese flip rates of 14.4\% and 14.2\%, respectively. DeepSeek and Gemini have lower English-vs.-Chinese flip rates at 11.1\%, but still change about one in nine judgments. Language-switched comparisons are also unstable, with flip rates from 10.7\% to 12.9\%. The Wilson intervals remain well above zero, with lower bounds from 8.8\% to 12.2\%.

\paragraph{Correctness changes nearly as often as preferences.}
The gold-correctness flip rates are close to the raw preference flip rates. This means that flips are not just harmless swaps between a correct answer and a tie. A language transformation often changes whether the judge agrees with the objective LLMBar label. Exact paired tests on strict correctness reject symmetry for every model-language comparison ($p \le 0.011$; Table~\ref{tab:sensitivity-significance}). For users relying on automatic judges to compare multilingual outputs, this is the more consequential failure mode.

\paragraph{No simple English preference in tie probes.}
The title asks whether the judge prefers English. The tie-probe results suggest a more nuanced answer: not systematically. Most translation-equivalent comparisons are judged as ties. DeepSeek reports ties 97.5\% of the time, GPT-4.1 Mini 95.6\%, Gemini 92.6\%, and Claude 82.3\%. When judges do not tie, Chinese wins more often than English for every model. The problem is therefore not a universal English-favoring prior; it is instability under language presentation.

\section{Extended Analysis}

\begin{table*}[t]
\centering
\small
\setlength{\tabcolsep}{6pt}
\renewcommand{\arraystretch}{1.08}
\begin{tabular}{lrrrrr}
\hline
Subset & \shortstack{EN\\acc. (\%)} & \shortstack{ZH\\acc. (\%)} & \shortstack{LS\\acc. (\%)} & \shortstack{EN--ZH\\judgment flip (\%)} & \shortstack{EN--LS\\judgment flip (\%)} \\
\hline
Adv-GPTInst & 85.9 & 82.5 & 81.9 & 11.1 & 10.5 \\
Adv-GPTOut & 74.5 & 68.9 & 71.5 & 16.5 & 15.7 \\
Adv-Manual & 80.6 & 74.7 & 76.5 & 14.4 & 10.1 \\
Adv-Neighbor & 80.6 & 77.0 & 75.8 & 11.8 & 12.3 \\
Natural & 92.7 & 87.9 & 88.6 & 12.9 & 12.3 \\
\hline
\end{tabular}
\caption{Subset-level averages across the four judges. Accuracy uses tie-half scoring; judgment flip compares transformed language conditions with English. All rates are percentages.}
\label{tab:source-aggregate}
\end{table*}

\begin{table}[t]
\centering
\small
\setlength{\tabcolsep}{4pt}
\renewcommand{\arraystretch}{1.08}
\begin{tabular}{lrrr}
\hline
Model & \shortstack{EN\\(\%)} & \shortstack{ZH\\(\%)} & \shortstack{LS\\(\%)} \\
\hline
Claude Haiku & 13.6 & 15.3 & 18.4 \\
DeepSeek & 9.3 & 11.5 & 12.7 \\
Gemini Flash & 10.3 & 16.0 & 22.2 \\
GPT-4.1 Mini & 15.0 & 16.7 & 19.1 \\
\hline
Mean & 12.1 & 14.9 & 18.1 \\
\hline
\end{tabular}
\caption{Position inconsistency by model and language condition. A case is inconsistent when the normalized winner changes after swapping answer order; all rates are percentages.}
\label{tab:position-extra}
\end{table}

\begin{table}[t]
\centering
\small
\setlength{\tabcolsep}{4pt}
\renewcommand{\arraystretch}{1.08}
\begin{tabular}{p{0.68\linewidth}rr}
\hline
Audit check & Count & \shortstack{Rate\\(\%)} \\
\hline
Unique transformed items & 419 & 100.0 \\
Complete EN/ZH/LS fields & 419 & 100.0 \\
Manual JSON repair & 1 & 0.2 \\
Mechanically flagged high-risk items & 19 & 4.5 \\
Items retained in sensitivity analysis & 400 & 95.5 \\
\hline
\end{tabular}
\caption{Automatic transformation audit. High-risk flags are conservative mechanical warnings based on empty fields, severe length shrinkage, or large numeric-token count changes; they are not treated as semantic labels.}
\label{tab:variant-audit}
\end{table}

\begin{table*}[t]
\centering
\small
\setlength{\tabcolsep}{6pt}
\renewcommand{\arraystretch}{1.08}
\begin{tabular}{llrrrr}
\hline
Model & \shortstack{Language\\pair} & N & \shortstack{Full judgment\\flip (\%)} & \shortstack{Filtered judgment\\flip (\%)} & \shortstack{Correctness\\test p} \\
\hline
Claude Haiku & EN--LS & 838 & 12.3 & 10.8 & 0.000131 \\
Claude Haiku & EN--ZH & 838 & 14.2 & 13.0 & 0.000598 \\
DeepSeek & EN--LS & 838 & 10.7 & 9.1 & 0.00804 \\
DeepSeek & EN--ZH & 838 & 11.1 & 9.8 & 0.0105 \\
Gemini Flash & EN--LS & 838 & 12.2 & 10.8 & 2.2e-05 \\
Gemini Flash & EN--ZH & 838 & 11.1 & 10.0 & 0.000313 \\
GPT-4.1 Mini & EN--LS & 838 & 12.9 & 12.2 & 1e-06 \\
GPT-4.1 Mini & EN--ZH & 838 & 14.4 & 13.8 & 2e-06 \\
\hline
\end{tabular}
\caption{Sensitivity and paired-significance checks. Filtered judgment flip excludes mechanically flagged high-risk transformation items. The p-value is an exact two-sided McNemar/binomial test on strict gold-correctness changes between English and the target language.}
\label{tab:sensitivity-significance}
\end{table*}

\begin{figure*}[t]
\centering
\includegraphics[width=0.47\linewidth]{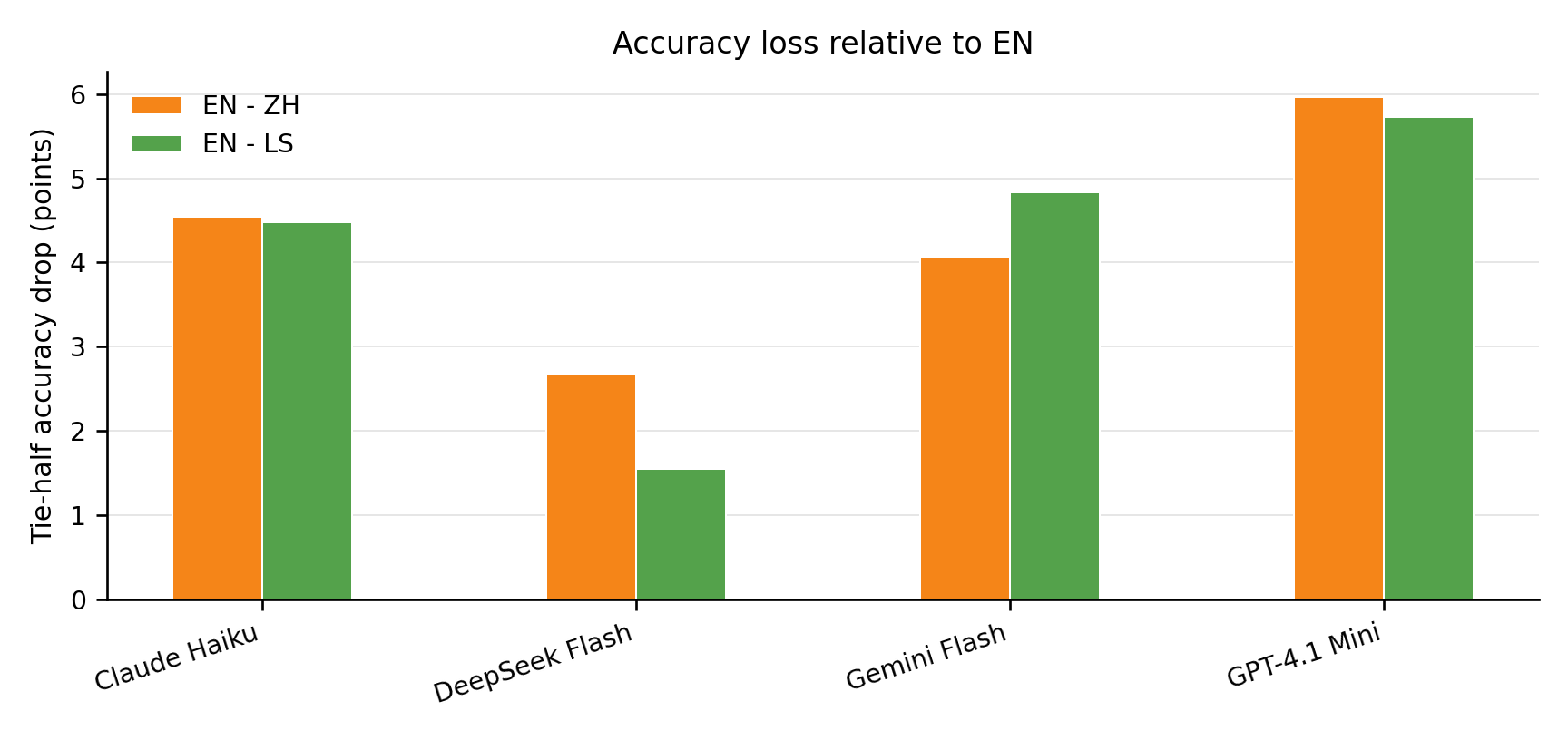}
\includegraphics[width=0.47\linewidth]{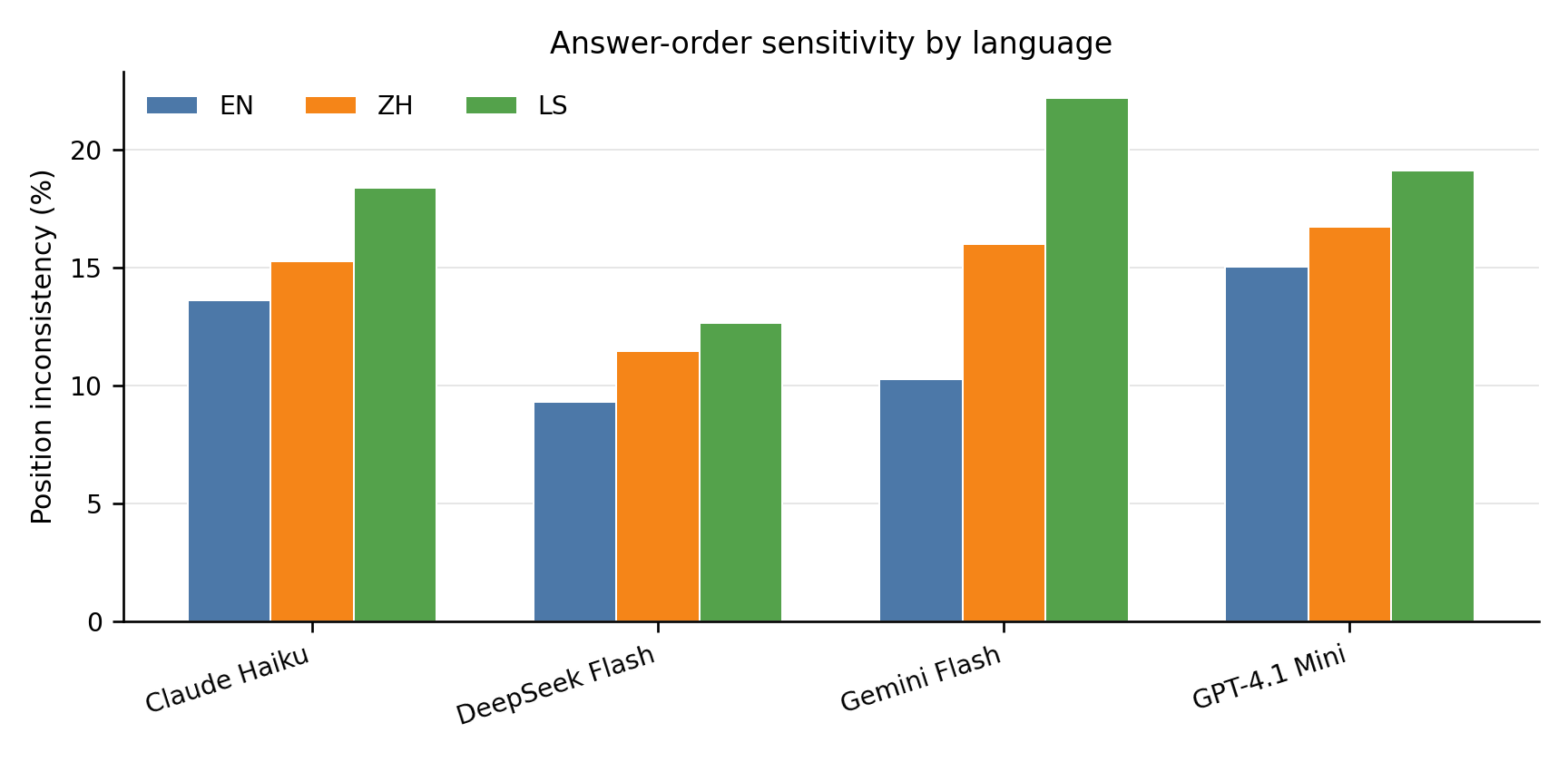}
\caption{Aggregate diagnostics. Left: tie-half accuracy loss relative to EN for ZH and LS. Right: answer-order inconsistency by language condition; a change after swapping A/B indicates position sensitivity.}
\label{fig:aggregate-extra}
\end{figure*}

\begin{figure*}[t]
\centering
\includegraphics[width=0.47\linewidth]{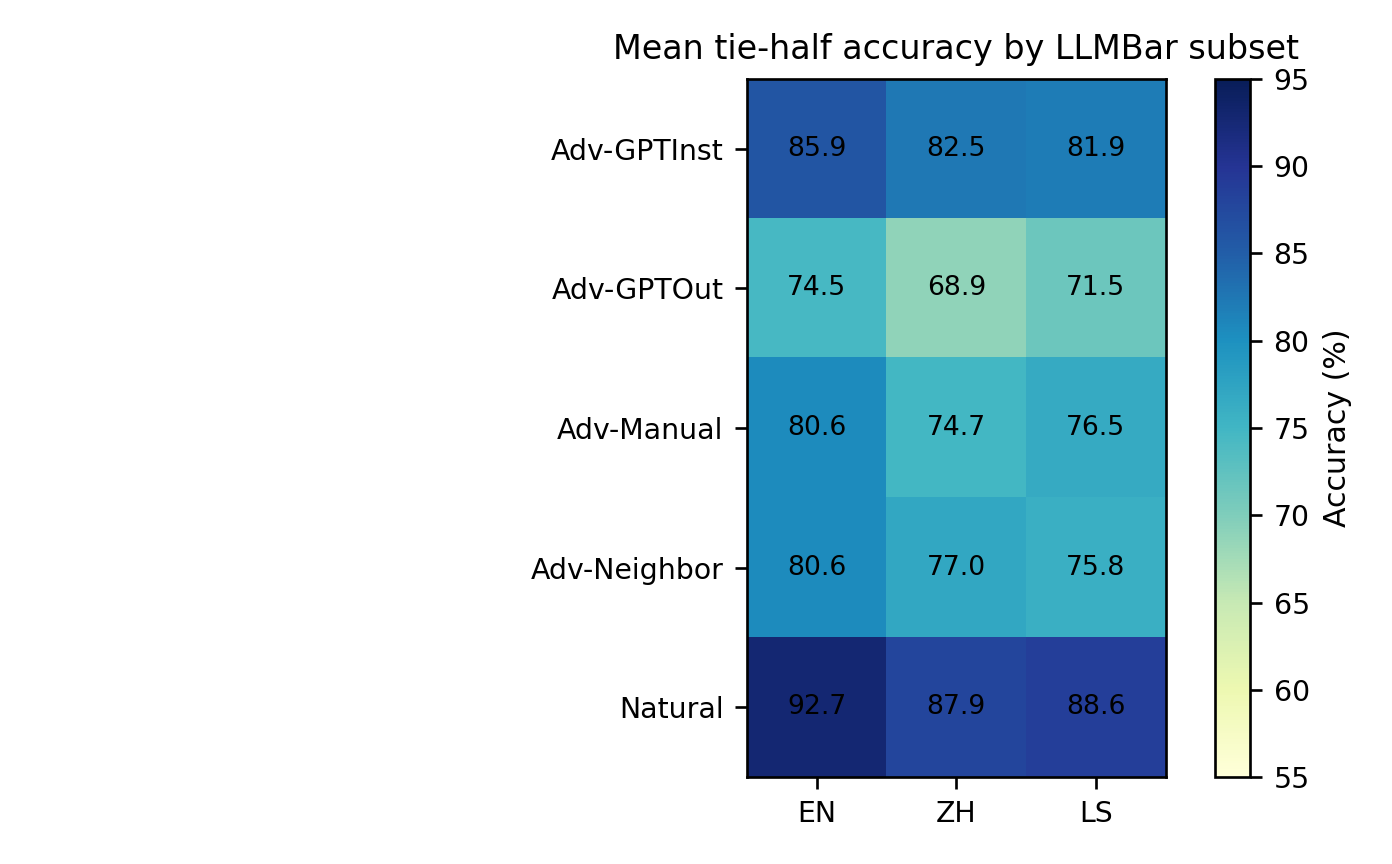}
\includegraphics[width=0.47\linewidth]{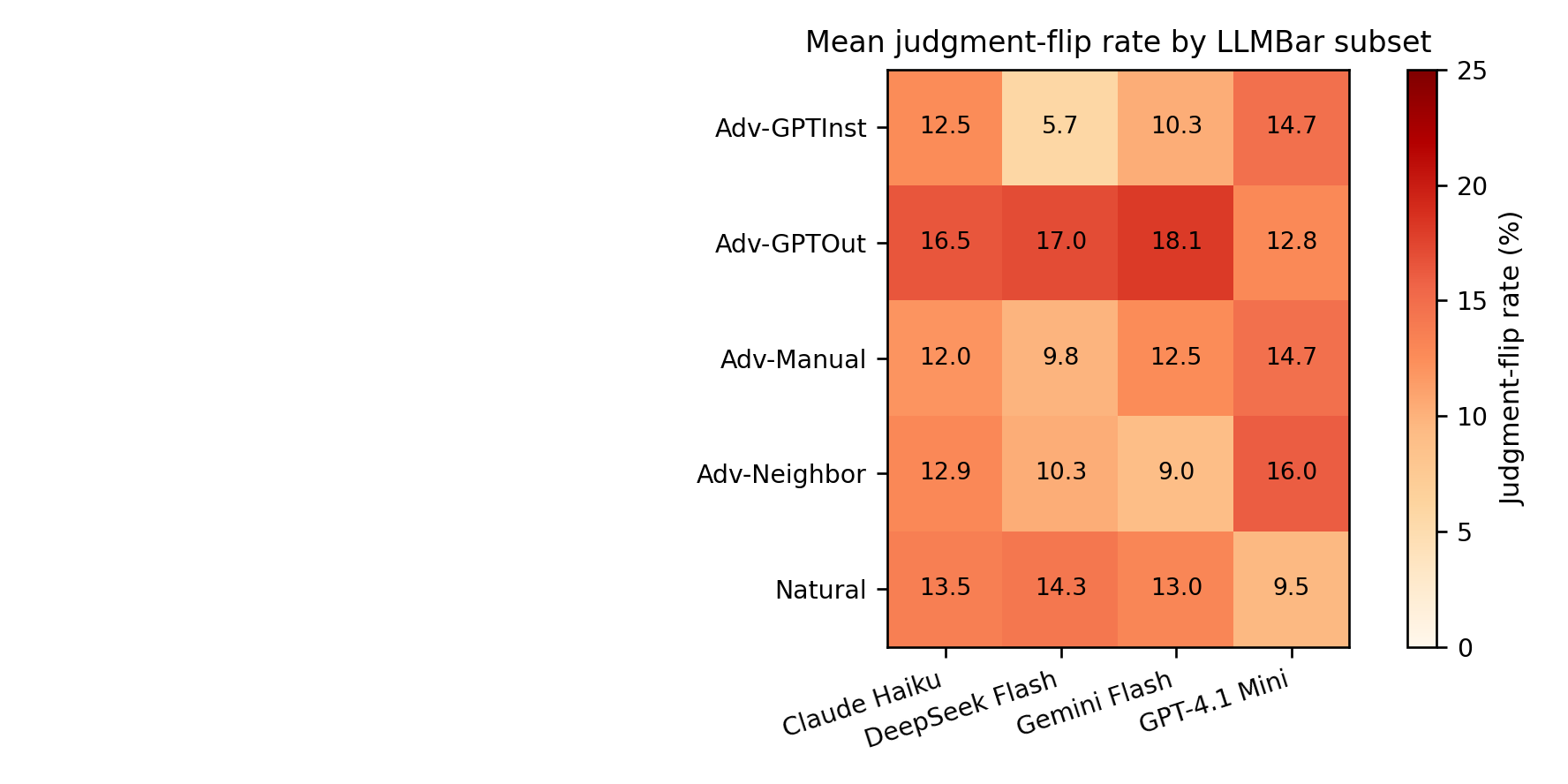}
\caption{Subset-level behavior on LLMBar. Left: mean tie-half accuracy by source subset and language condition. Right: mean judgment-flip rate by source subset and judge, averaged over EN--ZH and EN--LS.}
\label{fig:source}
\end{figure*}

\paragraph{Transformation audit supports a sensitivity check.}
Table~\ref{tab:variant-audit} shows that all 419 unique items have complete English, Chinese, and language-switched fields, with one manual JSON repair. The conservative mechanical audit flags 19 items (4.5\%) as high-risk because they contain severe length shrinkage or large numeric-token discrepancies. Excluding these items leaves 400 items and 12,800 quality judgments. The filtered flip rates in Table~\ref{tab:sensitivity-significance} remain substantial, ranging from 9.1\% to 13.8\%. This does not prove that every transformation is semantically perfect, but it reduces the chance that the main conclusion is driven only by a small set of visibly risky variants.

\paragraph{Subset averages reveal two distinct effects.}
Table~\ref{tab:source-aggregate} separates accuracy loss from decision instability. The Natural subset has the highest English accuracy at 92.7\%, but its English--Chinese flip rate is still 12.9\%. This means that high average accuracy does not guarantee stable judgments. Adv-GPTOut is the hardest subset, with mean English accuracy of 74.5\% and Chinese accuracy of 68.9\%, and it also has the largest average flip rates. The subset contains adversarially generated outputs that are often plausible but subtly wrong, so a translation or language-switched presentation can change which details the judge attends to.

\paragraph{Chinese and language-switching are not identical perturbations.}
Chinese translation and Chinese-English language-switching both reduce accuracy relative to English, but their failure patterns differ. Chinese creates larger drops on Adv-Manual and Adv-GPTOut, while language-switching creates the largest increase in position inconsistency. This difference is useful diagnostically. A pure translation condition tests whether semantic content survives a language change; a language-switched condition additionally tests whether the judge can handle mixed linguistic cues, partially translated task terms, and local changes in discourse structure. In real technical usage, language-switching is common: users may write the task in Chinese while keeping API names, mathematical expressions, software identifiers, or domain terms in English.

\paragraph{Language-switching amplifies position sensitivity.}
Figure~\ref{fig:aggregate-extra} and Table~\ref{tab:position-extra} show that language-switched inputs often increase answer-order sensitivity. GPT-4.1 Mini has 15.0\% position inconsistency in English, rising to 19.1\% in language-switched form. Claude rises from 13.6\% to 18.4\%, and Gemini rises from 10.3\% to 22.2\%. DeepSeek is more stable overall, but still rises from 9.3\% to 12.7\%. Averaged across models, position inconsistency is 12.1\% in English, 14.9\% in Chinese, and 18.1\% in language-switched form. This suggests that language-switching is not merely a language translation issue; it interacts with how pairwise prompts are parsed and compared.

\paragraph{Adversarial subsets are harder and less invariant.}
Figure~\ref{fig:source} shows that the Natural subset is easiest, while adversarial subsets are more difficult and less stable. Adv-GPTOut is particularly challenging: it has lower mean accuracy and higher language-flip rates across several models. This is consistent with LLMBar's goal of stressing evaluator reasoning. The language transformation appears to add another layer of difficulty on top of adversarial response construction.

\paragraph{Model families differ in failure mode.}
DeepSeek achieves the best average accuracy and lowest position inconsistency, but it also has high output-token usage because of reasoning content. Gemini achieves strong English accuracy but has relatively high language-switched position inconsistency and a high position-A rate. GPT-4.1 Mini and Claude Haiku are more sensitive to Chinese transformations. These differences indicate that a multilingual judge benchmark should report multiple diagnostics rather than only one aggregate accuracy.

\paragraph{The tie probe separates bias from instability.}
If a judge had a simple English preference, we would expect English to win many translation-equivalent probes. Instead, the non-tie decisions usually favor Chinese. This matters conceptually: language-related judge failures can occur even without a clear preference for a high-resource language. The judge may be inconsistent because translation changes perceived specificity, fluency, reasoning traces, or formatting rather than because it blindly rewards English.

\paragraph{Qualitative failure modes.}
Manual inspection of sampled disagreements suggests three recurring patterns. First, some judges overweight fluency after translation and become less strict about whether the response follows a hidden constraint in the instruction. Second, some language-switched prompts make an otherwise minor formatting mismatch more salient, especially when one answer keeps English terminology and the other is more fully localized. Third, adversarial examples often contain a strong but wrong answer; when the surface language changes, the judge may shift from checking the constraint to rewarding the more coherent response. These observations are not used as labels in the quantitative analysis, but they help explain why language flips are distributed across both correct-to-incorrect and incorrect-to-correct changes.

\paragraph{Reproducibility checks.}
The pipeline records every raw model response, parsed winner, confidence value, language condition, item ID, answer order, and token count when available. The analysis script deduplicates retries by stable judgment keys and reports only successful unique judgments. We also keep failed responses in the raw log during execution so that parser or endpoint problems can be separated from model behavior. This is important for API-based experiments because transient endpoint failures, model-specific formatting habits, and output-token limits can otherwise masquerade as evaluation results.

\section{Recommended Workflow}

Judge-LS is meant to be small enough for routine use. A practical evaluation workflow can be implemented in five stages. First, run the normal English judge benchmark with swapped answer order and report both strict and tie-half accuracy. Second, transform each item into the target deployment language and at least one realistic mixed-language condition. Third, audit the transformed items and run a sensitivity analysis that excludes flagged variants. Fourth, compute language-invariance flips against the English baseline, not only accuracy in each language, and attach confidence intervals or paired tests. Fifth, add translation-equivalent tie probes to distinguish language preference from general instability. If a model has high accuracy but high flip rate, it may still be unsuitable as the only judge for multilingual model selection.

This workflow does not require a new benchmark for every project. Any pairwise evaluation set with stable labels can be reused, provided that the transformation process preserves the response-quality relation. For deployment studies, the same idea can be applied to domain-specific tasks: customer-support answers, educational feedback, coding explanations, or safety-policy judgments. The key design principle is to perturb the language surface while keeping the intended preference label fixed.

\paragraph{Cost remains low.}
The expanded 419-item run required 13,408 unique judgments plus 419 language-variant generations. The observed or approximated usage is reported in Table~\ref{tab:usage}. Under reference public prices embedded in the script, the end-to-end run remains inexpensive for a small research project. The proxy endpoint may bill differently, so we report token counts separately from estimated cost.

\section{Discussion}

\paragraph{Implication for multilingual evaluation.}
A multilingual LLM benchmark often assumes that the judge's decision is comparable across language conditions. Judge-LS shows that this assumption should be tested. A 10--14\% flip rate is large enough to alter model rankings when systems have close scores. If a leaderboard uses an English judge prompt but evaluates Chinese or language-switched answers, the reported performance may mix answer quality with judge-language sensitivity.

\paragraph{Implication for prompt design.}
The prompts in this experiment explicitly tell the judge not to favor language, length, politeness, or formatting unless directly relevant. The observed instability therefore does not disappear with a simple instruction. More robust designs may need bilingual rubrics, explicit translation-equivalence checks, self-consistency over language variants, or cascades where a second judge adjudicates language-sensitive cases.

\paragraph{Relation to robustness evaluation.}
Judge-LS treats the judge as a system under test rather than as a transparent measuring device. This is closer to robustness evaluation than to ordinary scoring: the intended quality relation is held fixed, while the surface form is perturbed. A robust judge should preserve the core preference under such label-preserving transformations, just as robust classifiers should preserve decisions under irrelevant input perturbations. The language-flip and position-inconsistency metrics therefore measure whether the evaluator itself is stable enough to support multilingual model comparison.

\paragraph{Practical recommendation.}
For multilingual LLM-as-a-Judge studies, we recommend reporting at least five checks: answer-order swaps, translation or paraphrase invariance, tie probes for semantically equivalent answers, transformation-quality audits, and per-subset breakdowns with confidence intervals or paired tests. These checks are inexpensive compared with human annotation and can reveal whether an evaluation pipeline is brittle before it is used for model selection.

\section{Limitations}

First, the language variants are generated by an LLM. The automatic audit and filtered sensitivity analysis reduce obvious mechanical risks, but they do not replace bilingual human semantic validation; 19 items are flagged as high-risk and should be manually checked before using the variants as a release benchmark. Second, the tie probes use the gold answer because it gives a conservative high-quality equivalence condition; weaker-answer tie probes would further test error-preservation but require additional controls. Third, the study focuses on Chinese and Chinese-English language-switching. Other language pairs, scripts, and mixed-language communities may show different behavior. Fourth, only four API judge models are tested, and API versions can change. Fifth, LLMBar is pairwise and instruction-following oriented; tasks involving long-form reasoning, safety, medical advice, or multimodal grounding may require additional protocols.

\section{Conclusion}

We introduced Judge-LS, a lightweight protocol for evaluating whether LLM judges remain stable under English, Chinese, and Chinese-English language-switched presentations. On the full 419-item LLMBar benchmark, four API judges produced 13,408 successful judgments. All judges performed best in English, and language transformations caused 10.7--14.4\% preference flips. The result remains visible after excluding mechanically flagged high-risk transformations, and paired tests show significant strict-correctness changes. Yet translation-equivalent tie probes did not show a systematic English preference. The central risk is therefore not simply that judges prefer English; rather, judges can be sensitive to language presentation in ways that change correctness and interact with position bias. Judge-LS is inexpensive, training-free, and easy to extend, making it a practical diagnostic for multilingual evaluation pipelines.


\begin{thebibliography}{27}

\bibitem{fu2025reliable}
Xiyan Fu and Wei Liu.
\newblock 2025.
\newblock How Reliable is Multilingual LLM-as-a-Judge?
\newblock In \emph{Findings of the Association for Computational Linguistics: EMNLP 2025}, pages 11040--11053.

\bibitem{liu2023geval}
Yang Liu, Dan Iter, Yichong Xu, Shuohang Wang, Ruochen Xu, and Chenguang Zhu.
\newblock 2023.
\newblock G-Eval: NLG Evaluation using GPT-4 with Better Human Alignment.
\newblock In \emph{Proceedings of EMNLP 2023}, pages 2511--2522.

\bibitem{zeng2024llmbar}
Zhiyuan Zeng, Jiatong Yu, Tianyu Gao, Yu Meng, Tanya Goyal, and Danqi Chen.
\newblock 2024.
\newblock Evaluating Large Language Models at Evaluating Instruction Following.
\newblock In \emph{International Conference on Learning Representations}.

\bibitem{zheng2023judging}
Lianmin Zheng, Wei-Lin Chiang, Ying Sheng, Siyuan Zhuang, Zhanghao Wu, Yonghao Zhuang, Zi Lin, Zhuohan Li, Dacheng Li, Eric P. Xing, Hao Zhang, Joseph E. Gonzalez, and Ion Stoica.
\newblock 2023.
\newblock Judging LLM-as-a-Judge with MT-Bench and Chatbot Arena.
\newblock In \emph{Advances in Neural Information Processing Systems 36}.

\bibitem{papineni2002bleu}
Kishore Papineni, Salim Roukos, Todd Ward, and Wei-Jing Zhu.
\newblock 2002.
\newblock BLEU: A Method for Automatic Evaluation of Machine Translation.
\newblock In \emph{Proceedings of ACL 2002}, pages 311--318.

\bibitem{lin2004rouge}
Chin-Yew Lin.
\newblock 2004.
\newblock ROUGE: A Package for Automatic Evaluation of Summaries.
\newblock In \emph{Text Summarization Branches Out}, pages 74--81.

\bibitem{zhang2020bertscore}
Tianyi Zhang, Varsha Kishore, Felix Wu, Kilian Q. Weinberger, and Yoav Artzi.
\newblock 2020.
\newblock BERTScore: Evaluating Text Generation with BERT.
\newblock In \emph{International Conference on Learning Representations}.

\bibitem{rei2020comet}
Ricardo Rei, Craig Stewart, Ana C. Farinha, and Alon Lavie.
\newblock 2020.
\newblock COMET: A Neural Framework for MT Evaluation.
\newblock In \emph{Proceedings of EMNLP 2020}, pages 2685--2702.

\bibitem{dubois2023alpacafarm}
Yann Dubois, Xuechen Li, Rohan Taori, Tianyi Zhang, Ishaan Gulrajani, Jimmy Ba, Carlos Guestrin, Percy Liang, and Tatsunori B. Hashimoto.
\newblock 2023.
\newblock AlpacaFarm: A Simulation Framework for Methods that Learn from Human Feedback.
\newblock arXiv:2305.14387.

\bibitem{dubois2024lengthcontrolled}
Yann Dubois, Balazs Galambosi, Percy Liang, and Tatsunori B. Hashimoto.
\newblock 2024.
\newblock Length-Controlled AlpacaEval: A Simple Way to Debias Automatic Evaluators.
\newblock arXiv:2404.04475.

\bibitem{kim2024prometheus}
Seungone Kim, Jamin Shin, Yejin Cho, Joel Jang, Shayne Longpre, Hwaran Lee, Sangdoo Yun, Seonghyeon Shin, Sungdong Kim, James Thorne, and Minjoon Seo.
\newblock 2024.
\newblock Prometheus: Inducing Fine-grained Evaluation Capability in Language Models.
\newblock In \emph{International Conference on Learning Representations}.

\bibitem{wang2024fair}
Peiyi Wang, Lei Li, Liang Chen, Dawei Zhu, Binghuai Lin, Yunbo Cao, Qi Liu, Tianyu Liu, and Zhifang Sui.
\newblock 2024.
\newblock Large Language Models are not Fair Evaluators.
\newblock In \emph{Findings of the Association for Computational Linguistics: ACL 2024}.

\bibitem{panickssery2024selfpreference}
Nina Panickssery, Samuel R. Bowman, and Shi Feng.
\newblock 2024.
\newblock LLM Evaluators Recognize and Favor Their Own Generations.
\newblock In \emph{Advances in Neural Information Processing Systems 37}.

\bibitem{hu2020xtreme}
Junjie Hu, Sebastian Ruder, Aditya Siddhant, Graham Neubig, Orhan Firat, and Melvin Johnson.
\newblock 2020.
\newblock XTREME: A Massively Multilingual Multi-task Benchmark for Evaluating Cross-lingual Generalization.
\newblock In \emph{International Conference on Machine Learning}, pages 4411--4421.

\bibitem{bang2023multitask}
Yejin Bang, Samuel Cahyawijaya, Nayeon Lee, Wenliang Dai, Dan Su, Bryan Wilie, Holy Lovenia, Ziwei Ji, Tiezheng Yu, Willy Chung, et al.
\newblock 2023.
\newblock A Multitask, Multilingual, Multimodal Evaluation of ChatGPT on Reasoning, Hallucination, and Interactivity.
\newblock In \emph{Proceedings of IJCNLP-AACL 2023}, pages 675--718.

\bibitem{lai2023chatgpt}
Viet Dac Lai, Nghia Trung Ngo, Amir Pouran Ben Veyseh, Hieu Man, Franck Dernoncourt, Trung Bui, and Thien Huu Nguyen.
\newblock 2023.
\newblock ChatGPT Beyond English: Towards a Comprehensive Evaluation of Large Language Models in Multilingual Learning.
\newblock In \emph{Findings of the Association for Computational Linguistics: EMNLP 2023}, pages 13171--13189.

\bibitem{ahuja2023mega}
Kabir Ahuja, Harshita Diddee, Rishav Hada, Millicent Ochieng, Krithika Ramesh, Prachi Jain, Akshay Nambi, Tanuja Ganu, Sameer Segal, Maxamed Axmed, Kalika Bali, and Sunayana Sitaram.
\newblock 2023.
\newblock MEGA: Multilingual Evaluation of Generative AI.
\newblock In \emph{Proceedings of EMNLP 2023}, pages 4232--4267.

\bibitem{aibench2025}
Zicheng Zhang, Junying Wang, Yijin Guo, Farong Wen, Zijian Chen, Hanqing Wang, Wenzhe Li, Lu Sun, Yingjie Zhou, Jianbo Zhang, Bowen Yan, Ziheng Jia, Jiahao Xiao, Yuan Tian, Xiangyang Zhu, Kaiwei Zhang, Chunyi Li, Xiaohong Liu, Xiongkuo Min, Qi Jia, and Guangtao Zhai.
\newblock 2025.
\newblock AIBench: Towards trustworthy evaluation under the 45$^\circ$ law.
\newblock \emph{Displays}, page 103255.
\newblock doi: \href{https://doi.org/10.1016/j.displa.2025.103255}{10.1016/j.displa.2025.103255}.

\bibitem{zhang2025large}
Zicheng Zhang, Junying Wang, Farong Wen, Yijin Guo, Xiangyu Zhao, Xinyu Fang, Shengyuan Ding, Ziheng Jia, Jiahao Xiao, Ye Shen, Yushuo Zheng, Xiaorong Zhu, Yalun Wu, Ziheng Jiao, Wei Sun, Zijian Chen, Kaiwei Zhang, Kang Fu, Yuqin Cao, Ming Hu, Yue Zhou, Xuemei Zhou, Juntai Cao, Wei Zhou, Jinyu Cao, Ronghui Li, Donghao Zhou, Yuan Tian, Xiangyang Zhu, Chunyi Li, Haoning Wu, Xiaohong Liu, Junjun He, Yu Zhou, Hui Liu, Lin Zhang, Zesheng Wang, Huiyu Duan, Yingjie Zhou, Xiongkuo Min, Qi Jia, Dongzhan Zhou, Wenlong Zhang, Jiezhang Cao, Xue Yang, Junzhi Yu, Songyang Zhang, Haodong Duan, and Guangtao Zhai.
\newblock 2025.
\newblock Large Multimodal Models Evaluation: A Survey.
\newblock \emph{SCIENCE CHINA Information Sciences}, 68(12):221301--221369.
\newblock doi: \href{https://doi.org/10.1007/s11432-025-4676-4}{10.1007/s11432-025-4676-4}.

\bibitem{tian2019asd}
Yuan Tian, Xiongkuo Min, Guangtao Zhai, and Z. Gao.
\newblock 2019.
\newblock Video-based early ASD detection via temporal pyramid networks.
\newblock In \emph{IEEE International Conference on Multimedia and Expo}, pages 272--277.

\bibitem{tian2022ean}
Yuan Tian, Yichao Yan, Guangtao Zhai, G. Guo, and Z. Gao.
\newblock 2022.
\newblock EAN: Event adaptive network for enhanced action recognition.
\newblock \emph{International Journal of Computer Vision}, 130(10):2453--2471.

\bibitem{tian2024lowbitrate}
Yuan Tian, Guo Lu, Yichao Yan, Guangtao Zhai, L. Chen, and Z. Gao.
\newblock 2024.
\newblock A Coding Framework and Benchmark towards Low-Bitrate Video Understanding.
\newblock \emph{IEEE Transactions on Pattern Analysis and Machine Intelligence}.

\bibitem{tian2023nonsemantics}
Yuan Tian, Guo Lu, Guangtao Zhai, and Z. Gao.
\newblock 2023.
\newblock Non-semantics suppressed mask learning for unsupervised video semantic compression.
\newblock In \emph{Proceedings of the IEEE/CVF International Conference on Computer Vision}, pages 13610--13622.

\bibitem{tian2023clsa}
Yuan Tian, Yichao Yan, Guangtao Zhai, L. Chen, and Z. Gao.
\newblock 2023.
\newblock CLSA: A contrastive learning framework with selective aggregation for video rescaling.
\newblock \emph{IEEE Transactions on Image Processing}, 32:1300--1314.

\bibitem{tian2021selfconditioned}
Yuan Tian, Guo Lu, Xiongkuo Min, Z. Che, Guangtao Zhai, G. Guo, and Z. Gao.
\newblock 2021.
\newblock Self-conditioned probabilistic learning of video rescaling.
\newblock In \emph{Proceedings of the IEEE/CVF International Conference on Computer Vision}, pages 4490--4499.

\bibitem{tian2020motion}
Yuan Tian, Zhaohui Che, Wenbo Bao, Guangtao Zhai, and Z. Gao.
\newblock 2020.
\newblock Self-supervised Motion Representation via Scattering Local Motion Cues.
\newblock In \emph{European Conference on Computer Vision}, pages 71--89.

\bibitem{tian2025smcpp}
Yuan Tian, Xiaoyue Ling, Cong Geng, Q. Hu, Guo Lu, and Guangtao Zhai.
\newblock 2025.
\newblock SMC++: Masked learning of unsupervised video semantic compression.
\newblock \emph{IEEE Transactions on Pattern Analysis and Machine Intelligence}.

\end{thebibliography}
\end{document}